# A Transformer-Based Deep Learning Approach for Fairly Predicting Post-Liver Transplant Risk Factors

Can Li, MS[1], Xiaoqian Jiang, PhD[2] and Kai Zhang, PhD[2]

[1]Department of Biostatistics and Data Science, School of Public Health, The University of Texas Health Science Center at Houston, Houston, TX

[2]Department of Health Data Science and Artificial Intelligence, McWilliams School of Biomedical Informatics, The University of Texas Health Science Center at Houston, Houston, TX

Corresponding Author: Kai Zhang (Kai.Zhang.1@uth.tmc.edu)

**Abstract**

Liver transplantation is a life-saving procedure for patients with end-stage liver disease. There are two main challenges in liver transplant: finding the best matching patient for a donor and ensuring transplant equity among different subpopulations. The current MELD scoring system evaluates a patient's mortality risk if not receiving an organ within 90 days. However, the donor-patient matching should also consider post-transplant risk factors, such as cardiovascular disease, chronic rejection, etc., which are all common complications after transplant. Accurate prediction of these risk scores remains a significant challenge. In this study, we used predictive models to solve the above challenges. Specifically, we proposed a deep learning model to predict multiple risk factors after a liver transplant. By formulating it as a multi-task learning problem, the proposed deep neural network was trained to simultaneously predict the five post-transplant risks and achieve equal good performance by exploiting task-balancing techniques. We also proposed a novel fairness-achieving algorithm to ensure prediction fairness across different subpopulations. We used electronic health records of 160,360 liver transplant patients, including demographic information, clinical variables, and laboratory values, collected from the liver transplant records of the United States from 1987 to 2018. The model's performance was evaluated using various performance metrics such as AUROC and AUPRC. Our experiment results highlighted the success of our multitask model in achieving task balance while maintaining accuracy. The model significantly reduced the task discrepancy by 39%. Further application of the fairness-achieving algorithm substantially reduced fairness disparity among all sensitive attributes (gender, age group, and race/ethnicity) in each risk factor. It underlined the potency of integrating fairness considerations into the task-balancing framework, ensuring robust and fair predictions across multiple tasks and diverse demographic groups.

## 1. Introduction

Liver transplantation is a critical treatment option for individuals with severe liver diseases. Although it provides a life-saving treatment for patients with end-stage liver disease, there remain concerns about potential mortality and post-transplant complications. Liver transplant post-transplant risk prediction is an important area of research, and machine learning is increasingly important in developing predictive models.

Several studies have investigated the use of machine learning algorithms for predicting post-transplant survival and identifying high-risk patients. For example, Nitski et al. noted that liver transplant recipients have an increased risk of cancer beyond the first year after transplantation, and deep learning models outperformed predicting 1-year and 5-year causes of death due to cardiovascular causes, infection, graft failure, or cancer [1]. Kantidakis et al. identified retransplantation as the strongest predictor of graft failure or death using Cox and machine learning models to predict the post-transplant survival rates [2].

Our work represents the first attempt to predict multiple risk factors for liver transplant patients simultaneously. Previous studies used simple clinical scoring systems like the MELD (Model for End-Stage Liver Disease) score or built models for individual risk prediction. For instance, the MELD score only predicts 90-day mortality if the patient is not receiving a liver. However, the success of a liver transplant goes beyond patient survival; it also depends on post-transplant risks like cardiovascular disease, rejection, diabetes, etc. These risk factors are crucial when making decisions to match patients to donors.

The current strategy of independent prediction models for each risk factor lacks a standardized methodology for ensuring equal performance among these models. Consequently, some risk predictions underperform others because this approach does not adequately leverage the interrelationships among the risks. Research demonstrated that post-transplant risk factors, such as malignancy, diabetes, rejection, infection, and cardiovascular complications, are highly interdependent [3], indicating that a joint prediction approach would be more effective than single risk predictions. Multi-task prediction models have demonstrated remarkable performance when the tasks have correlations [4]. However, the effectiveness of multi-task learning depends on the availability and quality of large datasets with existing methods limited by data scarcity and overfitting risks. In this work, we proposed using multi-task balancing techniques to jointly predict the multiple clinically essential risk factors (malignancy, diabetes, rejection, infection, and cardiovascular complications) after liver transplantation and achieved high prediction performance.

Additionally, liver transplantation faces the challenge of ensuring fairness across different subpopulations due to the limited organ donations. Numerous studies have highlighted regional and demographic disparities in liver transplantation access [5,6]. For example, due to geographic variation in liver transplant access, the supply-to-patient demand ratio differs across the country, causing significant regional differences in liver transplantation rates [7]. Furthermore, demographic disparities exist in liver transplant access. Before the MELD era, black patients had lower transplant rates than white patients [8]. In the MELD era, the disparity in liver transplant rates between females and males has increased [9]. Addressing these disparities is crucial to ensure equitable access to liver transplantation for all individuals.

Efforts have been undertaken to address issues of fairness and disparity in liver transplant access. To balance efficiency and fairness in organ allocation, the OPTN/UNOS (Organ Procurement and Transplantation Network/United Network for Organ Sharing) has implemented policies [10]. Nonetheless, further endeavors are necessary to ensure equitable access to liver transplantation for all individuals. Risk prediction systems are particularly important as they can be applied in real-world settings for effective donor-patient matching. Therefore, we propose a fairness-achieving algorithm to ensure equal prediction performances on different subpopulations defined by sensitive attributes, such as gender, race, ethnicity, and age group.

Our proposed model utilized a large liver transplant database, including comprehensive donor and recipient demographics, vital signs, lab test results, diagnosis, and treatment data. It utilized both donors' and recipients' biomarkers to predict post-transplant risks, allowing for a more accurate prediction of donor-patient matching based on various biomarkers. The complex underlying biological process determines the goodness of matching between donor and patient.

Therefore, the innovations of our study are:
- The proposed model integrates donors' and recipients' biomarkers to improve the donor-recipient matching system.
- We develop an accurate multi-task prediction model with transformer-based neural networks and task-balancing techniques to achieve equal performance for multiple risk predictions.

- A fairness-achieving algorithm is proposed to ensure equal prediction performances on different subpopulations defined by sensitive attributes.

## 2. Related works

Predicting the post-transplant risks for liver transplant is important because it can better guide the patient screening process for best organ match. Meanwhile, achieving equal prediction performance across different subpopulations is essential to ensure fairness.

**Machine learning and deep learning models:** Lau et al. emphasized that the ability to predict graft failure predictions or primary nonfunction at the time of liver transplant is essential for optimizing the utilization of limited donor liver resources. In this regard, machine learning algorithms have demonstrated notable effectiveness [11]. Additionally, Liu et al. applied machine learning to predict short-term survival following liver transplantation, enabling the early identification of high-risk patients [12]. Furthermore, in the post-transplant survival prediction, Kantidakis et al. highlighted that machine learning models, utilizing gradient boosting, random forests, and decision trees, outperformed traditional Cox models, thereby drawing a significant comparison between traditional statistical methods and advanced machine learning techniques [2]. Ferrarese et al. have also developed machine-learning models to predict pretransplant survival among liver transplant patients. The goal of these models is to aid in the optimization of organ allocation and reduce waitlist mortality [13]. Nitski et al. demonstrated that deep learning models, including the Transformer, Temporal Convolutional Network, Recurrent Neural Network, and Multilayer Perceptron, can incorporate longitudinal information to predict long-term outcomes after liver transplantation. Furthermore, they also compared these models to logistic regression, focusing on their capacity to predict post-transplantation complications resulting in death across various timeframes [1].

**Fairness in predictive modeling:** Bias can be introduced during data collection or model training in machine learning [14]. Therefore, it becomes essential to consider fairness during the prediction. Fairness algorithms in machine learning can be categorized as pre-processing, in-processing, or post-processing [15]. Pre-processing algorithms focus on removing bias before the model training process. Calmon et al. introduced a convex optimization method to reduce discrimination in data preprocessing [16]. Post-processing algorithms adjust model predictions after training in fairness criteria. Pleiss et al. minimized error disparity among different populations by calibrating equalized odds [17]. The majority of efforts in achieving fairness are focused on in-processing algorithms, which integrate fairness into model training by including fairness metrics as regularization terms for optimization [18]. Ding et al. employed knowledge distillation with a two-step fair machine learning framework targeting graft failure prediction in liver transplants [19]. Li et al. developed a multi-task learning strategy with a dynamic weighting scheme to improve fairness in predictive models. This approach overcomes the constraints of non-differentiable fairness metrics by dynamically adjusting the gradients of various prediction tasks during neural network back-propagation [20]. A recent framework called MDANN, based on adversarial learning, demonstrated fair predictions on biomedical imaging datasets by effectively addressing bias across multiple sensitive features through its adversarial modules within the framework [21].

**Multi-task learning:** Multi-task learning (MTL) trains a model on multiple related tasks simultaneously, enabling the model to learn robust and transferable representations by integrating commonalities and differences between the tasks. MTL often outperforms single-task learning on related tasks, minimizing multiple loss functions concurrently. However, balancing learning rates is challenging, as one task can dominate and degrade others due to gradient conflict [22]. Imbalanced data distributions between tasks, caused by differing sample sizes for each task's training data, impede MTL's performance. Techniques like focal loss can help counteract this issue of class imbalance between tasks [23]. Lee and Son proposed gradient-based meta-learning to balance multi-task networks by separating shared and task-specific layers. Updating the shared layer with a single gradient step and inner/outer loop training mitigates gradient imbalance between tasks. This approach boosts generalization and efficiency while achieving state-of-the-art performance on computer vision MTL benchmarks [24]. Navon et al. introduced Nash-MTL, a multi-task learning optimization procedure that views gradient combination as a bargaining game between tasks to reach a *Nash Bargaining Solution* for balancing learning rates and data distributions across tasks [25]. The CoD-MTL framework employs multitask learning to jointly analyze post-liver transplant causes of death, specifically focusing on rejection and infection [26].

## 3. Methodology

The transformer model has become extremely popular and revolutionized natural language processing, leading to significant advancements in machine translation and text summarization [27]. The TabTransformer, a variation of the Transformer model, uses the self-attention mechanism to capture complex interactions between features in tabular data [28]. It allows for handling high cardinality categorical variables efficiently through embeddings. In this study, we aim to use TabTransformer to predict multiple post-transplant risks simultaneously.

### 3.1. The TabTransformer multi-task learning model

We propose a novel multi-task TabTransformer deep learning model for predicting five risk factors in post-liver transplant. The model consists of a column embedding, layer normalization, a stack of Transformer layers, and a multilayer perceptron. We extend the TabTransformer architecture for multi-task learning by adapting the output layers for each risk prediction.

The multi-task TabTransformer model handles related tasks simultaneously in a joint prediction framework, sharing knowledge across the tasks to improve generalization and efficiency and capturing complex interactions and dependencies.

The architecture of our TabTransformer Multi-task Fairness-achieving prediction model is shown in **Figure 1**.

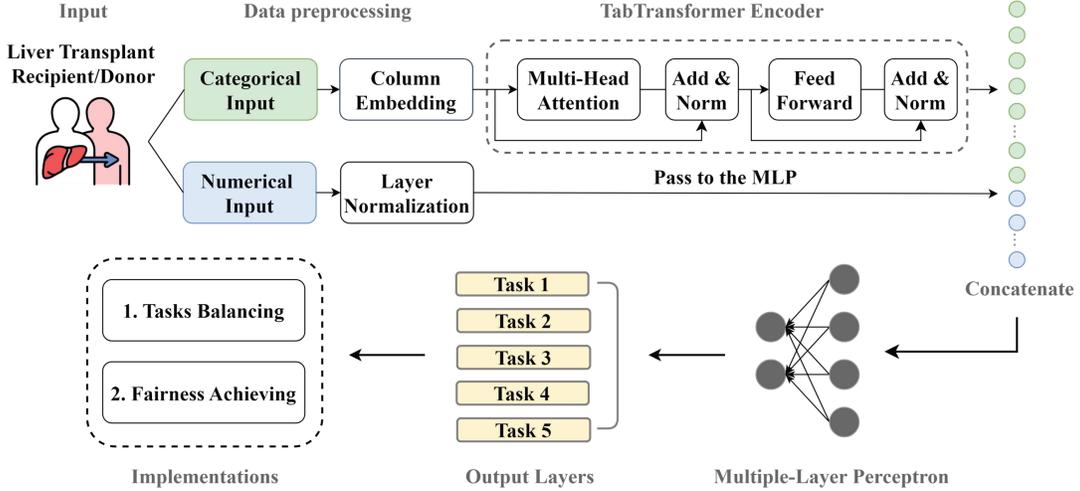

**Figure 1**. Multiple risk fairness-achieving prediction model architecture.

Let $x_{cat}$ denotes all the categorical features as $x_{cat} = \{x_1, x_2, \ldots, x_p\}$, where $p$ represents the number of categorical features. Variable $x_{cont}$ denotes all the numerical features in $R^q$ where $q$ represents the number of numerical features. In the tabular data, the combined feature set $X$ would be the union of the continuous and categorical features, i.e., $X = x_{cat} \cup x_{cont}$. We use $D = \{X, y_1, y_2, \ldots, y_M\}$ to denote a dataset where each $y_m$, for $m$ in $\{1, \ldots, M\}$, represents the set of output labels for the $m^{th}$ task, and $M$ is the number of tasks. In this paper, we have five tasks as the corresponding risks. We consider the binary classification for each task in $\{y_1, y_2, \ldots, y_M\}$.

### 3.1.1. Column Embedding and Layer Normalization

We propose a method for generating embeddings of categorical features, which we term as $E_\phi(x_{cat})$. $E_\phi(x_{cat})$ consists of a set of embeddings $\{e_{\phi_1}(x_1), \ldots, e_{\phi_p}(x_p)\}$ corresponding to each categorical feature. Each feature $x_j$, where $j \in \{1, \ldots p\}$, is converted into a parametric metric embedding space of dimensionality $d$ using *column embedding*. To elucidate, each categorical column, $j$ links to an embedding lookup table, denoted as $e_{\phi_j}$. For a feature $j$ with $d_j$ classes, its embedding table $e_{\phi_j}(.)$ has $(d_j + 1)$ embeddings, with the extra one representing a missing value.

For each continuous variable $x_{cont} \in R^q$ in our neural network, we do the layer normalization as $LN_q = \gamma_q \odot N_q + \beta_q$, where $\gamma_q$ denotes the learnable scaling and $\beta_q$ denotes the shifting parameters [29].

### 3.1.2. Multiple Layer Perceptron (MLP) as Multi-task Output Layers

The parametric embeddings $E_\phi(x_{cat})$ serve as input to the initial Transformer layer. The output from this first layer then serves as the input to the second Transformer layer, and the process continues. Parametric embeddings are incrementally transformed into contextual embedding through the final Transformer layer, aggregating context from other embeddings. The sequence of Transformer layer function $f_\theta$ processes the parametric embeddings $\{e_{\phi_1}(x_1), \ldots, e_{\phi_p}(x_p)\}$ and generates corresponding contextual embeddings $\{h_1, \ldots, h_p\}$ where $h_p \in R^d$.

Following this, categorical features generated contextual embeddings $\{h_1, \ldots, h_p\}$ are concatenated with continuous features $x_{cont}$ to produce a vector of dimension $(d \times p + q)$. This vector is then fed into a Multilayer Perceptron (MLP), denoted as $g_\psi^m$ for each task $m$ and predicts the target $y^m$.

In the multi-task model, the loss function $L_m(X, y^m)$ is minimized to train all the parameters of the TabTransformer end-to-end, using first-order gradient methods. The parameters include $\phi$ for column embedding, $\theta$ for Transformer layers, and $\psi$ for the MLP layer. We implement five distinct prediction networks, each structured as an MLP. Each MLP takes the concatenated vector and outputs the logit of the probability. For this multi-task setting, the model adopts a strategy of hard parameter sharing.

The overall loss function for the multi-task model considering all five tasks can be defined as the sum of individual task losses as:

$$Loss(X,Y) = \sum_{m=1}^{M} L_m(X, y^m) \tag{1}$$

where $L_m(X, y^m)$ is the loss function for each task $m$ defined as $BCE(g_\psi^m(f_\theta(E_\phi(x_{cat})), x_{cont}), y^m)$. Here, binary cross-entropy (BCE) is the loss function for this binary classification problem.

### 3.1.3. TabTransformer Encoder

The TabTransformer model includes multiple Transformer layers, each featuring a series of components: a multi-head self-attention mechanism, Layer Normalization, and a position-wise feed-forward network [28]. These components process the input features and generate refined feature representations for the subsequent layers. This mechanism uses three parameter matrices for each input embedding: $Key$ $(K)$, $Query$ $(Q)$, and $Value$ $(V)$, all of which reside in different dimensions. Suppose there are $p$ input embeddings corresponding to $p$ categorical features. In this case, the $Key$ and $Query$ matrices can be represented as $K \in R^{p \times k}$ and $Q \in R^{p \times k}$ respectively, where $k$ is the dimension of the key and query vectors. The $Value$ matrix can be denoted as $V \in R^{p \times v}$, with $v$ being the dimension of the value vector.

These embeddings are then passed through the self-attention mechanism, which computes the attention weights between each pair of embeddings. This is done using the equation $A = softmax((QK^T)/\sqrt{k})$, where $A \in R^{p \times p}$ quantifies the attention weights. These weights serve to highlight the relative importance of each input embedding.

Subsequently, the attention weights are used to transform the $Value$ matrix, producing contextualized representations of the original input features. This is achieved by operating: $Attention\ (K, Q, V) = A \cdot V$.

Finally, the output of the attention head, which has dimension $v$, is mapped back to the original embedding dimension $d$ using a fully connected layer. This final output, a transformed version of the original input features, is fed into subsequent layers for predicting tasks [28].

**3.2. Task balancing method**

Given the imbalance across the various tasks, it is crucial to incorporate task-balancing techniques into our approach. We propose the concept of task prioritization, which is guided by the model's task-specific prediction performance. In previous work, we utilized dynamic re-weighting to promote fairness across subgroups defined by sensitive attributes such as gender, race, and age. The key part was to adjust the weighting factor $w_m(t)$ on the task $m$ at epoch $t$ using the performance metric on these tasks at the previous epoch $(t-1)$. The performance metric can be flexibly chosen based on the specific evaluation metric to be optimized [20]. In this work, the modified loss functions at epoch $t$ are the weighted sum of the loss of each task $m$,

$$Loss_{task-balancing}(t) = \sum_{m=1}^{M} \left(1 - w_m^{metric}(t-1)\right)^{\alpha} L_m(t) \qquad (2)$$

where $L_m(t)$ is the loss function of the task $m$ at epoch $t$, and *(metric)* is the evaluation metric chosen to be optimized. The scheme achieves task balance by adjusting the magnitudes of each task's weight based on the previous training epoch's performance $w_m^{metric}(t-1))$. At each epoch $t$, the algorithm reduces the task's weight, which performs better on the metric than the average and then focuses more on the other task, and vice versa. The parameter $\alpha$ is a hyperparameter that adjusts the magnitude of the weight, therefore determining the strength of weighting. This method aims to achieve task balancing during model training to address the imbalance issue arising from the dataset. Unlike prior work, this paper innovatively adapts the dynamic re-weighting mechanism to the novel problem of task balancing. Rather than repurposing the existing approach, we demonstrate its adaptability and expand its application to a new domain.

In our AUROC-Weighted model, a dynamic re-weighting strategy is employed where evaluation metrics can be flexibly chosen depending on the specific evaluation metric to be balanced across different tasks. For this study, we specifically utilize the Area Under the Curve of the Receiver Operating Characteristic (AUROC). The scores obtained from AUROC assessments will then be used to create a distinct dynamic re-weighting scheme. The scheme re-weights the gradients during back-propagation for each task. This approach ensures that tasks with lower performance receive more focus during the training process, thereby improving the overall balance and effectiveness of the model across all tasks. We could also utilize other metrics like the Area Under Precision-Recall Curve (AUPRC), accuracy, precision, or recall for the re-weighting scheme.

### 3.3. Fairness achieving method

To promote fairness across various demographic categories, we present a new disparity measure, $D_m$. This is calculated as the disparity between the prediction for the most advantaged subgroup and the overall average prediction for all subgroups.

The sensitive attribute (such as age, gender, or race/ethnicity) splits the entire population into disjoint subgroups. For example, the subgroup can be Male/Female (when the sensitive attribute is gender); White/Black/Asian/etc. (when the sensitive attribute is race/ethnicity); etc. We denote the subgroups defined by the sensitive attribute as a set $A = \{1, \dots, |A|\}$. The calculations are tailored for each sensitive attribute and are specific to each task $m$. In this context, the $i$ in the equation represents any subgroup within the range $\{1, \dots, |A|\}$. The mathematical representation of $D_m$ is:

$$D_m = \max_{i \in A}(F_i^m) - E_{i \in A}[F_i^m] \tag{3}$$

where $F_i^m$ is the fairness metric of interest. This measure can be flexibly chosen based on the specific fairness metric to be optimized. For example, the metric is the positive prediction rate when *demographic parity* is optimized; the average of the true positive and false positive rate when *equalized odds* is optimized, etc.

The loss of fairness is defined as the summation of the $D_m$ on all tasks,

$$Loss_{fairness}(t) = \sum_{m=1}^{M} D_m \tag{4}$$

In this strategy, we incorporated the fairness metrics widely used to evaluate the fairness prediction, including but not limited to demographic parity and equalized odds, to reduce disparity among different subgroups within each sensitive attribute. In the multi-task setting, for each task $m \in [1, \ldots, M]$, $\hat{Y}$ represents the predictions, $Y$ represents the true label.

Demographic parity is the equality of the predicted positive rate among all subgroups defined by the sensitive attribute.

$$\mathbb{P}(\hat{Y} = 1 | A = i, Task = m) = \mathbb{P}(\hat{Y} = 1 | A = j, Task = m), \forall i, j \in A \tag{5}$$

When demographic parity is the fairness metric that is to be optimized among subgroups, the per-task fairness metric on group $i$, i.e., $F_i^m$, is defined as,

$$F_i^m \triangleq \mathbb{P}(\hat{Y} = 1 | A = i, Task = m) \tag{6}$$

Similarly, equalized odds is another widely used fairness criterion, which requires equal true positive rates and equal false positive rates across all subgroups,

$$\begin{aligned}
\mathbb{P}(\hat{Y} = 1 | Y = 1, A = i, Task = m) &= \mathbb{P}(\hat{Y} = 1 | Y = 1, A = j, Task = m), \forall i, j \in A, \\
\mathbb{P}(\hat{Y} = 1 | Y = 0, A = i, Task = m) &= \mathbb{P}(\hat{Y} = 1 | Y = 0, A = j, Task = m), \forall i, j \in A.
\end{aligned} \tag{7}$$

The $F_i^m$ is defined as the average of true positive rate and false positive rate,

$$F_i^m \triangleq \frac{\mathbb{P}(\hat{Y} = 1 | Y = 1, A = i, Task = m) + \mathbb{P}(\hat{Y} = 1 | Y = 0, A = i, Task = m)}{2} \tag{8}$$

The $F_i^m$ can be empirically estimated within a batch (denoted as $\hat{F}_i^m$) of data during model training.

### 3.4. Balancing Tasks Fairly: A Unified Approach

We propose our end-to-end algorithm, which accomplishes two objectives simultaneously -- task balancing and achieving fairness during model training, see **Algorithm 1.** The algorithm works in general for a wide variety of task-balancing metrics and fairness evaluation metrics. Therefore, we use 'metric' to denote the general performance metric to be chosen for balancing the tasks, and $\hat{F}_i^m(t)$ can also be defined as the 'fairness metric' of choice to evaluate the inequity. The goal of the proposed method is to ensure the prediction model achieves equal accurate prediction on all tasks while maintaining fairness among different subgroups simultaneously (gender, age group, race/ethnicity).

The unified approach operates in two steps: 1) Task balancing is performed first based on the selected 'metric' to mitigate task imbalance. 2) Next, fairness is promoted between subgroups within each task using the chosen' fairness metric'. By applying fairness optimization after initial task balancing, the approach can mitigate biases related to sensitive attributes in each task.

In subsequent sections, we detail our discussion in two distinct parts. Firstly, we introduce a multi-task model focused solely on task balancing using the $Loss_{task-balancing}(t)$. This model is referred to as the 'Multi-task AUROC-Weighted model'. Following that, our second model integrates task-balancing with fairness metrics into the total loss: $Loss_{task-balancing}(t) + Loss_{fairness}(t)$. Depending on the fairness metric employed, this model is designated as 'Multi-task AUROC-Weighted + DP' when demographic parity is the chosen metric, and 'Multi-task AUROC-Weighted + EO' when applying equalized odds.

### 3.5 Comparison Study

We assessed various methodologies within the TabTransformer model to compare each task prediction performance and discrepancies.

**Single Task Approach:** Our foundational method involved applying binary prediction for individual risk factors. The loss function for a single task can be defined as the BCE loss associated with that task represented as $L_m(X, y^m)$.

**Single Task with Focal Loss** [30]: To improve regular single-task learning, we incorporated focal loss to address the class imbalance problem within each individual task. It focuses on distinguishing between positive and negative classes in a more nuanced way. The loss function for a single task incorporating focal loss can be formulated as

$$L_m^{FL}(X, y^m) = -\alpha_{balanced}(1 - prob_{balanced})^{\gamma} \log(prob_{balanced}) \quad (9)$$

Here, $prob_{balanced} = prob \times y^m + (1 - prob) \times (1 - y^m)$ represents the adjusted predicted probability for the given task. $\alpha_{balanced} = \alpha \times y^m + (1 - \alpha) \times (1 - y^m)$ is the balanced weighting factor. $\gamma$ is the tunable focusing parameter larger or equal to 0.

**Multi-task balancing** [24]: To provide a comprehensive comparison, we also implemented the single gradient step update method for task balancing, a strategy proposed by Lee and Son.

Initially, this method computes the loss for each task. Subsequently, it updates the shared layer parameters with a single gradient descent step, keeping the task-specific layers constant. This update is mathematically captured by the equation:

$$h_m^{share} \leftarrow h^{share} - \beta \nabla_{h^{share}} L_m\left(f(h^{share}, h_m^{task})\right) \quad (10)$$

After this, the method recalculates the loss for each task, but this time using the updated shared layers $h_m^{share}$ and the original task-specific layer $h_m^{task}$. The shared layers are then further refined using the sum of the losses from all tasks, represented by the formula:

$$h^{share} \leftarrow h^{share} - \eta \nabla_{h^{share}} \sum_{m=1}^{M} L_m(f(h_m^{share}, h_m^{task})) \quad (11)$$

where $\beta$ and $\eta$ are the step sizes. Finally, the task-specific layers are updated based on the equation:

$$h_m^{task} \leftarrow h_m^{task} - \beta \nabla_{h_m^{task}} L_m(f(h^{share}, h_m^{task})) \quad (12)$$

This method offers an approach to balance tasks in multi-task learning by emphasizing separate updates for shared and task-specific layers using gradient-based techniques.

| | |
|---|---|
| **Algorithm 1.** Task-balancing Incorporating Fairness-achieving | |
| 1 | **Inputs** $D = \{X, y_1, y_2, \ldots, y_M\}$ |
| 2 | **Define** the tasks $\{1, \ldots, M\}$ |
| 3 | **Define** sensitive attributes (age, gender, race/ethnicity) and their subgroups $\{1, \ldots, |A|\}$ |
| 4 | **Define** the evaluation metric for task-balancing, e.g., AUROC, AUPRC, accuracy, precision, |
| 5 | **Define** the fairness metric for fairness-achieving, e.g., demographic parity (DP), equalized |
| 6 | **Initialized** weights $w_m(t)$ for all tasks $m$ |
| 7 | **Initialized** $\hat{F}_i^m(t)$ for all tasks $m$ and subgroup $i$ |
| 8 | **for** each epoch $t$ in number_epochs do |
| 9 |   **for** each batch $b$ in batches do |
| 10 |     **for** each task $m$ in $M$ do |
| 11 |     // Task balancing using chosen evaluation metric (AUROC or other metrics) |
| 12 |       calculate $L_m(t)$ |
| 13 |       calculate $w_m^{metric}(t-1))$ |
| 14 |     **end for** |
| 15 |     // Compute the sum of losses for task-balancing across all tasks |
| 16 |     $Loss_{task-balancing}(t) = \sum_{m=1}^{M} \left(1 - w_m^{metric}(t-1)\right)^{\alpha} L_m(t)$ |
| 17 |     // Calculate the predicted positive rate for chosen fairness metric for each subgroup in |
| 18 |     **for** each task $m$ in $M$ do |
| 19 |       **for** each subgroup $i$ in $A$ do |
| 20 |         calculate $\hat{F}_i^m$ of epoch $t$ for batch data $b$ |
| 21 |       **end for** |
| 22 |     **end for** |
| 23 |     // Compute disparity among all subgroups in a specific attribute |
| 24 |     $D_m = \max(\hat{F}_i^m) - E_{i \in A}[\hat{F}_i^m], E_{i \in A}[\hat{F}^m] = 1/|A| \sum_{i=1}^{|A|} \hat{F}_i^m$ |
| 25 |     // Compute the sum of losses for fairness-achieving across all tasks |
| 26 |     $Loss_{fairness}(t) = \sum_{m=1}^{M} D_m$ |
| 27 |     // Sum of the task-balancing loss and fairness-achieving loss |
| 28 |     $Loss(t) = Loss_{task-balancing}(t) + Loss_{fairness}(t)$ |
| 29 |   **end for** |
| 30 | **end for** |

# 4. Experiment Setting

## 4.1. Dataset

We collected a large cohort of 160,360 post-liver transplants patients using data drawn from the Organ Procurement and Transplantation Network (OPTN), maintained by the United Network for Organ Sharing (UNOS), from 1987 to 2018 [31]. It covers a broad range of information, including transplantations and waiting list registrations, demographic profiles, lab results, transplant events, and follow-up data. We used 5-fold cross-validation in our experiments. The data was split of 60% for training, 20% for validation, and 20% as a holdout test set. We report the average test performance across all folds with standard deviations to provide comprehensive insights into the results.

Our study particularly focuses on five risk factors (malignancy, diabetes, rejection, infection, and cardiovascular complications) that significantly influence post-liver transplant outcomes in **Figure 2**. The bar plot illustrates the prevalence of the five risk factors for post-liver transplant outcomes within the dataset. It clearly shows an imbalance in the distribution with rejection accounting for up to 23.77% while the least represented risk factor only constitutes 4.10%. This uneven distribution presents a considerable challenge when predicting multi-tasks concurrently, underscoring the necessity of considering task balance during our model development.

The proportion of each risk factor among post-liver transplant cases varied unevenly over the years, as shown in **Figure 3**. The proportion of rejection risk notably declined, while diabetes displayed a prominent peaked pattern, reaching its maximum around 2005 before decreasing. Similarly, the proportion of malignancy cases showed a more moderate peaked trend. In contrast, the proportion of infections declined more gradually over time. Meanwhile, cardiovascular risks remained relatively stable as a proportion of post-liver transplant cases over the years.

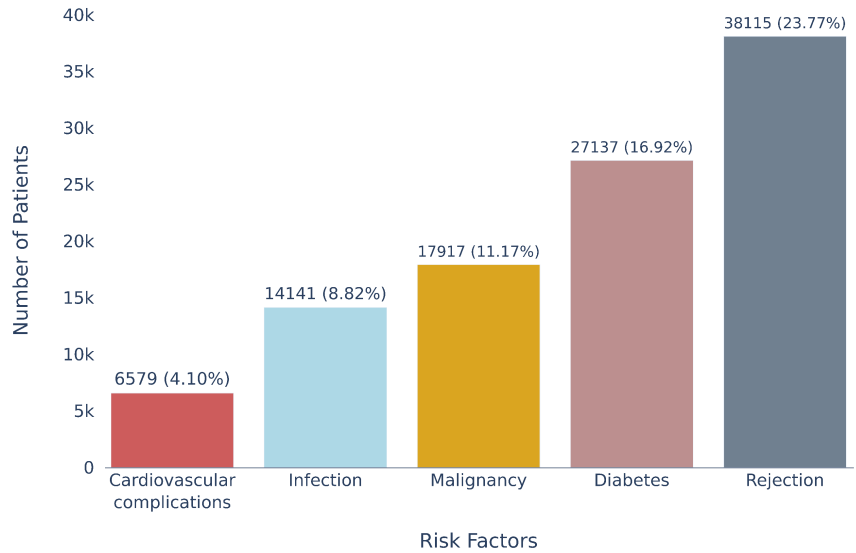

**Figure 2.** Architecture of the Multiple Risks Prediction Model for Post-Liver Transplant Outcomes. Patient counts for each risk factor are sequenced in ascending order, starting with cardiovascular complications, infection, malignancy, diabetes, and rejection.

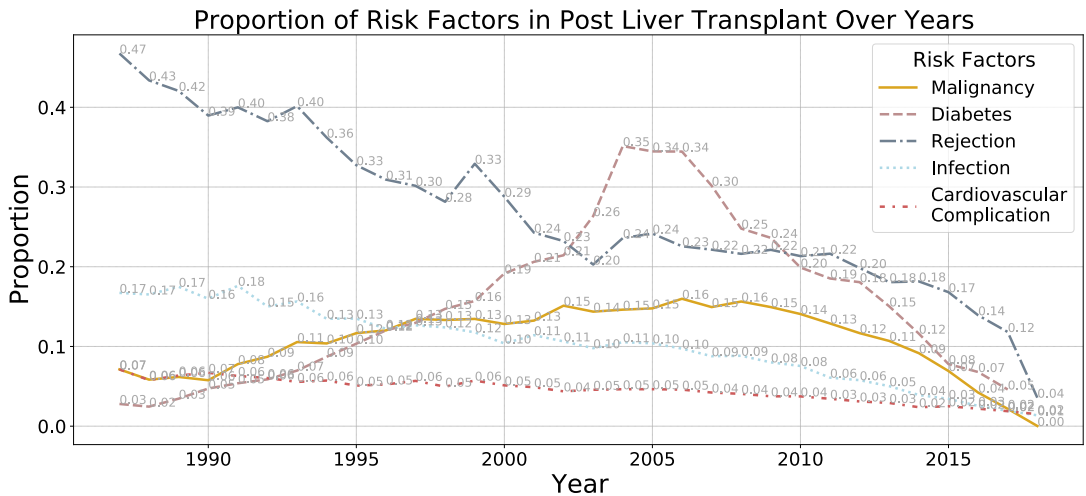

**Figure 3**. Imbalance of Risk Factors Proportions in Post-Liver Transplant Cases Over Years. Malignancy: Displays a moderate peaked trend. Diabetes: indicates an increase, peaking in 2005, followed by a decline. Rejection: Denotes a steady decrease in its proportion over the years. Infection: Represents a slow, gradual decline in its incidence. Cardiovascular: Remains consistently stable over the observed period.

An overview of the patient's demographic information in the liver transplant dataset is provided in **Table 1**. It includes various aspects such as age, gender, ethnicity, race, and education level, which helps identify potential disparities in the liver transplant population.

**Table 1**. Liver Transplant Patient Demographic Information

| | |
|---|---|
| **Total** | 160,360 (100%) |
| **Age** | |
| <15 | 15,135 (9.44%) |
| 15-29 | 8,208 (5.12%) |
| 30-44 | 23,056 (14.38%) |
| 45-59 | 73,704 (45.96%) |
| 60-74 | 40,005 (24.95%) |
| >=75 | 252 (0.16%) |
| **Gender** | |
| Male | 100,624 (62.75%) |
| Female | 59,736 (37.25%) |
| **Ethnicity/Race** | |
| White | 115,783 (72.20%) |
| Black | 15,180 (9.47%) |
| Asian | 6,458 (4.03%) |
| Hispanic | 21,028 (13.11%) |
| Other | 1,911 (1.19%) |
| **Education** | |
| None | 604 (0.38%) |
| Grade School (0-8) | 8,129 (5.07%) |
| High School (9-12) | 48,799 (30.43%) |
| Attended College | 26,124 (16.29%) |
| Associate/bachelor's degree | 17,873 (11.15%) |
| Post-college Graduate Degree | 7,552 (4.71%) |
| N/A (<5 years old) | 9,105 (5.68%) |
| Unknown | 42,174 (26.30%) |

Our study draws attention to disparities in post-liver transplant risk factors across subgroups in each sensitive attribute (gender, age, and race/ethnicity), depicted in **Figure 4**. For example, pediatric patients had a much higher rejection rate of 39.13% compared to 21.97% in adults. Meanwhile, adult patients display higher percentages of other risk factors. For instance, the diabetes rate among adult patients is 18.39%, while the corresponding rate for pediatric patients is 4.43%.

Race and ethnicity further reveal significant disparities among different subgroups. For instance, white patients encounter a 12.81% malignancy rate, markedly higher than those observed in other racial and ethnic subgroups. Regarding gender, the differences in risk factors, except for rejection, appear more balanced than other sensitive attributes. The rejection is a disease with the highest disparity compared to other risk factors. The disparity percentages are 6.04%, 17.16%, and 9.77% among gender, age, and race, respectively.

Such disparities underscore the urgency to address inequalities and tailor interventions to meet the specific needs of diverse subgroups in post-liver transplant care.

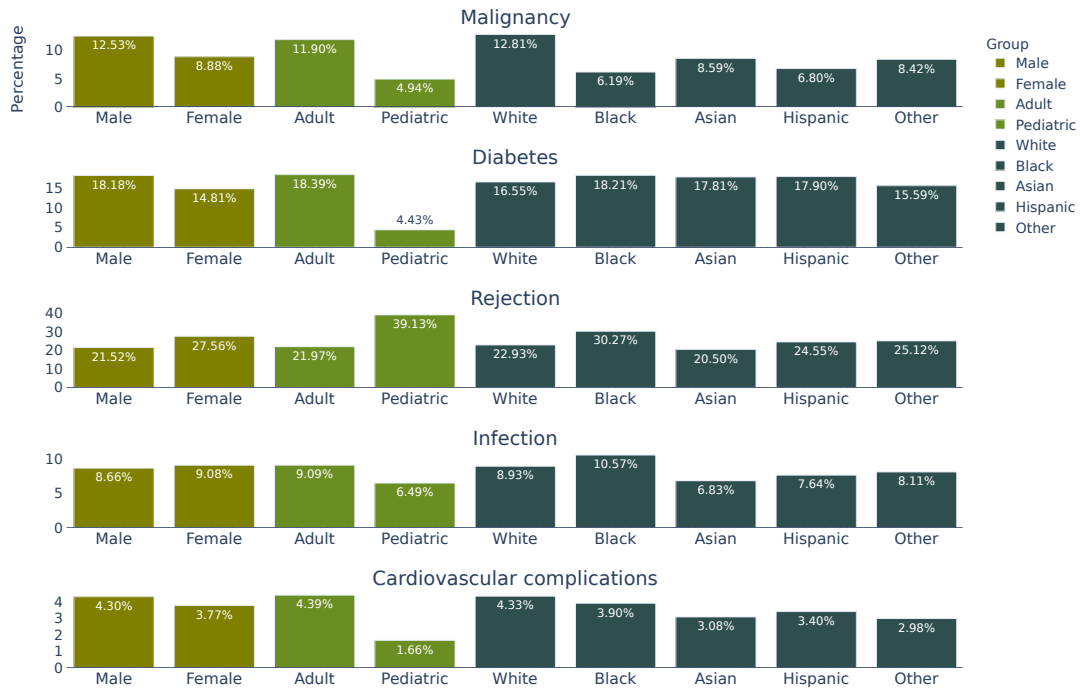

**Figure 4.** The disparity in subgroups of sensitive attributes across five risk factors post-liver transplant. Displayed as a bar plot, the figure details the occurrence rates of five risks - malignancy, diabetes, rejection, infection, and cardiovascular complications - for each demographic subgroup. The sensitive attributes are broken down into three main categories: Gender (Male, Female), Age Group (Adult, Pediatric), and Race/Ethnicity (White, Black, Asian, Hispanic, and Other). Each row presents a specific risk, comparing its prevalence among the subgroups of the three sensitive attributes.

### 4.2. Features selection and extraction

The post-transplant risks are largely decided by the quality of matching between the donor and the recipient, which is essentially a latent complex biological process. Therefore, we selected a comprehensive set of 52 features from patients and additional 65 features from donors to capture biological compatibility. The features are detailed in **Table S.1** and **Table S.2.** Unlike the MELD system, which primarily matches patient and donor based on blood body size, age, geographical considerations, and four blood tests [32], our model can learn patterns from historical data to optimize donor-recipient marching with extensive features.

The proposed model will guide donor-recipient matching for waiting list patients by predicting the risk scores of five risks before the transplantation happens. It can be used as a patient screening tool once an organ is available. For this purpose, we only select features from the data recorded before the liver transplant, including demographic information and laboratory test results. The pre-transplant features offered valuable insights into the patient's overall health status before the transplant, which is crucial for understanding the prediction of the selected risk factors.

### 4.3. Evaluation Metrics

The model's performance was assessed based on two primary objectives: (I) achieving task balancing across five tasks; and (II) ensuring fairness between subgroups in each sensitive attribute.

For objective (I), we evaluate the performances of different tasks and compare the maximum performance discrepancy among tasks before and after our task balancing algorithm. The performance metrics used to evaluate task balancing included AUROC and AUPRC. For objective (II), we compared the performance of the balanced multi-task model before and after applying the fairness-achieving algorithm. The fairness metrics we examined comprised demographic parity difference and equalized odds difference.

The definitions of these metrics are as follows, $\hat{Y}$: Prediction; $Y$: True label; $A$: The subgroups in each sensitive attribute, $m$: Specific task.

*Demographic Parity Difference:*

$$max_{i,j \in A}[P(\hat{Y} = 1 | A = i, Task = m) - P(\hat{Y} = 1 | A = j, Task = m)] \quad (13)$$

*Equalized Odds Difference:*

$$\begin{aligned}
max_{i,j \in A} &[P(\hat{Y} = 1 | Y = 1, A = i, Task = m) \\
&- P(\hat{Y} = 1 | Y = 1, A = j, Task = m)] \\
+ max_{i,j \in A} &[P(\hat{Y} = 1 | Y = 0, A = i, Task = m) \\
&- P(\hat{Y} = 1 | Y = 0, A = j, Task = m)]
\end{aligned} \quad (14)$$

## 5. Results

### 5.1. Evaluate the effectiveness of task-balancing

Our research included an ablation study to evaluate the effectiveness of different methodologies on task balancing and their performance. We compared the proposed Multi-task AUROC-Weighted model against single-task binary prediction, single-task with focal loss for class balance, multi-task without task-balancing, and multi-task learning with a single gradient step update. As evidenced in **Table 2**, the proposed model effectively improved the task balancing performance across the five tasks. For consistent evaluation, all approaches used the same TabTransformer architecture and hyperparameters.

Analysis of the results for Cardiovascular Complications (Task 5) revealed salient performance differences. The single-task had AUROC and AUPRC of 0.6148 (±0.0220) and 0.0598 (±0.0039). In contrast, the multi-task without task-balancing showed superior values of 0.6572 (±0.0084) and 0.0732 (±0.0057). Our AUROC-Weighted model championed exceptional task balancing and the best accuracy, with AUROC of 0.6600 (±0.0097) and AUPRC of 0.0753 (±0.0081). The table's final row highlights these variations by recording the difference between maximum and minimum AUROC or AUPRC across the five tasks for each method. Our AUROC-Weighted model minimized task disparities, with 0.0715 AUROC and 0.3150 AUPRC variation. Compared to the single-task AUROC disparity of 0.1172, we achieved a 39% reduction in disparity. This reduction was calculated using the formula (initial disparity-reduced disparity)/initial disparity.

In conclusion, the results highlight an important insight: multi-task learning inherently utilizes shared representations between related tasks, which enables it to outperform single-task predictions as evidenced by the higher scores. Additionally, task balancing optimizes each task and handles varied data distributions better than class balancing. Importantly, this model strikes an optimal balance—it proficiently narrows disparities with only a slight compromise on the accuracy of predictions.

**Table 2.** Prediction performance of the task-balancing model (Task 1: Malignancy, Task 2: Diabetes, Task 3: Rejection, Task 4: Infection, Task 5: Cardiovascular Complications). Each entry presents the average AUROC and AUPRC results over 5-fold for each method, along with standard deviations. The last row shows the difference in performance metrics among all five tasks of the task-balancing model. On the other hand, the methodology with the smallest AUROC and AUPRC disparity across tasks is marked bold with a star.

|  | Single-task | | Single-task (Focal Loss) | | Multi-task | | Multi-task[24] | | Multi-task AUROC-weighted | |
|---|---|---|---|---|---|---|---|---|---|---|
|  | AUROC | AUPRC | AUROC | AUPRC | AUROC | AUPRC | AUROC | AUPRC | AUROC | AUPRC |
| Task 1 | 0.6932 (±0.0068) | 0.2103 (±0.0032) | 0.6911 (±0.0115) | 0.2087 (±0.0094) | 0.7046 (±0.0066) | 0.2195 (±0.0073) | 0.7029 (±0.0073) | 0.2149 (±0.0077) | 0.7031 (±0.0047) | 0.2175 (±0.0059) |
| Task 2 | 0.7325 (±0.0053) | 0.3342 (±0.0095) | 0.7331 (±0.0049) | 0.3363 (±0.0092) | 0.7398 (±0.0044) | 0.3439 (±0.0093) | 0.7392 (±0.008) | 0.3411 (±0.0135) | 0.7315 (±0.0066) | 0.3336 (±0.0106) |
| Task 3 | 0.6633 (±0.0047) | 0.3790 (±0.0062) | 0.6664 (±0.0053) | 0.3845 (±0.0116) | 0.6710 (±0.0026) | 0.3958 (±0.0048) | 0.6708 (±0.0044) | 0.3952 (±0.0071) | 0.6683 (±0.0034) | 0.3903 (±0.0058) |
| Task 4 | 0.7015 (±0.0073) | 0.1826 (±0.0059) | 0.6984 (±0.0113) | 0.1801 (±0.0075) | 0.7117 (±0.0082) | 0.1923 (±0.0050) | 0.7113 (±0.0089) | 0.1928 (±0.0080) | 0.7115 (±0.0099) | 0.1926 (±0.0072) |
| Task 5 | 0.6148 (±0.0220) | 0.0598 (±0.0039) | 0.6162 (±0.0097) | 0.0599 (±0.0036) | 0.6572 (±0.0084) | 0.0732 (±0.0057) | 0.6571 (±0.0075) | 0.0714 (±0.0040) | 0.6600 (±0.0097) | 0.0753 (±0.0081) |
| Diff | 0.1176 | 0.3192 | 0.1169 | 0.3246 | 0.0826 | 0.3226 | 0.0821 | 0.3238 | **0.0715*** | **0.3150*** |

### 5.2. Evaluate the effectiveness of the fairness-achieving algorithm

After conducting the task-balancing method using our Multi-task AUROC-Weighted model, we then added the fairness achievement techniques. We performed a comparative analysis between the multi-task AUROC-Weighted model and their fairness-enhanced counterparts. It was achieved by incorporating our proposed fairness-achieving algorithm (using either Demographic Parity or Equalized Odds). The comparison utilized demographic parity difference and equalized odds difference as key evaluation metrics, as detailed in **Table 3.**

In our experiments with the Multi-task AUROC-Weighted model, we identified pronounced disparities in rejection risk (Task 3) across different gender, age, and racial groups. Notably, the risk of rejection demonstrated a significant Demographic Parity difference (DPD) of 0.1642 (±0.0051) and an Equalized Odds Difference (EOD) of

0.3058 (±0.0099), particularly highlighting disparities between the Adult and Pediatric subgroups. However, introducing fairness metrics - Demographic Parity and Equalized Odds - into the model mitigated these disparities. With the Multi-task AUROC-Weighted + DP model, the DPD for Rejection (Task 3) was lowered to 0.0039 (±0.0009), and the EOD reduced to 0.0222 (±0.0018). On the other hand, implementing the Multi-task AUROC-Weighted + EO resulted in a DPD of 0.0107 (±0.0016) and an EOD of 0.0134 (±0.0038) for the same task. Additionally, for risks with minimal inherent disparities across gender, age, and race, our fairness-enhanced models still decreased discrepancies. Infection (Task 4) and Cardiovascular Complications (Task 5) saw substantial reductions when processed with the proposed fairness methodologies, demonstrating robust disparity minimization.

The cumulative density functions (CDFs) of predicted probabilities for all five risks are depicted in **Figure 5** across various subgroups for each sensitive attribute. CDFs allow for a comprehensive view of how predicted risks were distributed within these subgroups. For example, for rejection risk (Task 3), the Multi-task AUROC-Weighted model's CDFs showed differing proportions of adults versus pediatric patients at each risk level. At a predicted risk of 0.25, about 70% of adults but only 10% of pediatric patients fall at or below this level. In contrast, with the proposed fairness model, the CDFs aligned more closely - about 60% of both subgroups are predicted to have a rejection risk of 0.25 or less. This alignment indicates more equitable prediction across subgroups. Similarly, the proposed model consistently aligns CDF curves between subgroups across other sensitive attributes, demonstrating its robustness in achieving fairness. To quantify gaps between subgroup CDFs, we annotated Demographic Parity Difference (DPD) values, where values closer to 0 indicate fairer predictions across subgroups. The proposed model's low DPD values validate its effectiveness in delivering equitable risk predictions across diverse demographics.

Overall, the proposed fairness-achieving method significantly reduced the disparity in demographic parity and equalized odds among subgroups of gender, age, and race/ethnicity on each task. This achievement demonstrates the effectiveness of our proposed solution in achieving fairness among different subgroups.

**Table 3**. We compared the fairness performance of the Multi-task AUROC-Weighted model against two fairness-enhanced models: the Multi-task AUROC-Weighted + DP and the Multi-task AUROC-Weighted + EO. Comparison of fairness across gender, age group, and race/ethnicity. Demographic Parity Difference (DPD) and Equalized Odds Difference (EOD) were the evaluation metrics employed. Each value describes the maximum disparity on a certain metric among different subgroups (Male vs. Female; Adult vs. Pediatric; White/Black/Asian/Hispanic/Other). We highlighted the instances where our fairness-achieving models demonstrated improved performance compared to the Multi-task AUROC-Weighted model.

| | | Multi-task AUROC-Weighted | | Multi-task AUROC-Weighted + DP | | Multi-task AUROC-Weighted + EO | |
|---|---|---|---|---|---|---|---|
| | | DPD | EOD | DPD | EOD | DPD | EOD |
| Age | Task 1 | 0.0903 (±0.0040) | 0.2142 (±0.0110) | **0.0033 (±0.0008)** | **0.0234 (±0.0015)** | **0.0009 (±0.0006)** | **0.0172 (±0.0041)** |
| | Task 2 | 0.1172 (±0.0078) | 0.2486 (±0.0166) | **0.0017 (±0.0008)** | **0.0162 (±0.0026)** | **0.0031 (±0.0007)** | **0.0108 (±0.0025)** |
| | Task 3 | 0.1642 (±0.0051) | 0.3058 (±0.0099) | **0.0039 (±0.0009)** | **0.0222 (±0.0018)** | **0.0107 (±0.0016)** | **0.0134 (±0.0038)** |
| | Task 4 | 0.0243 (±0.0025) | 0.0476 (±0.0106) | **0.0022 (±0.0010)** | **0.0117 (±0.0039)** | **0.0012 (±0.0011)** | **0.0090 (±0.0054)** |
| | Task 5 | 0.0891 (±0.0062) | 0.1975 (±0.0164) | **0.0041 (±0.0012)** | **0.0125 (±0.0064)** | **0.0010 (±0.0005)** | **0.0122 (±0.0072)** |
| Gender | Task 1 | 0.0537 (±0.0030) | 0.1131 (±0.0061) | **0.0009 (±0.0006)** | **0.0062 (±0.0021)** | **0.0035 (±0.0010)** | **0.0031 (±0.0014)** |
| | Task 2 | 0.0348 (±0.0015) | 0.0659 (±0.0042) | **0.0010 (±0.0004)** | **0.0089 (±0.0030)** | **0.0029 (±0.0011)** | **0.0050 (±0.0029)** |
| | Task 3 | 0.0577 (±0.0032) | 0.1136 (±0.0076) | **0.0015 (±0.0012)** | **0.0028 (±0.0014)** | **0.0047 (±0.0013)** | **0.0044 (±0.0031)** |
| | Task 4 | 0.0119 (±0.0016) | 0.0222 (±0.0074) | **0.0007 (±0.0004)** | **0.0055 (±0.0039)** | **0.0013 (±0.0009)** | **0.0053 (±0.0041)** |
| | Task 5 | 0.0159 (±0.0030) | 0.0333 (±0.0075) | **0.0008 (±0.0008)** | **0.0039 (±0.0038)** | **0.0011 (±0.0008)** | **0.0037 (±0.0022)** |
| Race/ Ethnicity | Task 1 | 0.0899 (±0.0065) | 0.1885 (±0.012) | **0.0085 (±0.0034)** | **0.0184 (±0.0051)** | **0.0089 (±0.004)** | **0.0200 (±0.0080)** |
| | Task 2 | 0.0230 (±0.0051) | 0.0488 (±0.0102) | **0.0083 (±0.0029)** | **0.0158 (±0.0061)** | **0.0065 (±0.0027)** | **0.0146 (±0.0051)** |
| | Task 3 | 0.0717 (±0.0043) | 0.1422 (±0.0119) | **0.0095 (±0.0028)** | **0.0183 (±0.0058)** | **0.0113 (±0.0027)** | **0.0240 (±0.0080)** |
| | Task 4 | 0.0518 (±0.0046) | 0.1049 (±0.0134) | **0.0098 (±0.0029)** | **0.0184 (±0.006)** | **0.0114 (±0.0034)** | **0.0241 (±0.0127)** |
| | Task 5 | 0.0384 (±0.0073) | 0.0855 (±0.0115) | **0.0091 (±0.0028)** | **0.0216 (±0.0052)** | **0.0077 (±0.0037)** | **0.0270 (±0.0096)** |

### 5.3. Evaluate feature importance

Following task balancing, we sought to assess potential unfairness introduced by the proposed Multi-task AUROC-Weighted model in the prediction scheme. To achieve this, we analyzed feature importance within the model. We utilized the permutation-based feature importance evaluation technique, shuffled feature values amongst patients, and examined the corresponding model performance changes. This technique allowed us to identify features significantly impacting the model's performance.

The top five features that most critically impact the prediction of each post-liver transplant risk factor are presented in **Figure 6a**. These predominant features have a higher relative impact on the performance of AUROC. For instance, the latest calculated BMI results (END_BMI_CALC) are essential in predicting diabetes risk, while the recipient ascites at transplant (ASCITES_TX) aids in predicting rejections. It is important to note that the proposed Multi-task AUROC-Weighted model relies heavily on demographic features such as age group, race/ethnicity, and gender for predictions rather than on biomarkers. For example, the age group (AGE_GROUP) feature is most important in predicting cardiovascular complications. Race/ethnicity (ETHACT) is the top feature of malignancy and rejection. It indicates that predictive models may rely on inappropriate features for prediction without enforcing fairness constraints, resulting in unfair outcomes.

In **Figure 6b** and **Figure 6c**, we implemented the AUROC-weighted task balancing method, incorporating either the Demographic Parity (DP) or the Equalized Odds (EO), respectively. Both models were designed to mitigate the impact of the age group feature on the model's performance. We observed that the age group feature no longer appears among the top five important features for any risk factors, demonstrating the efficacy of our fairness interventions.

Furthermore, we employed these same fairness-achieving methods on two other sensitive attributes: gender and race/ethnicity. The impacts of these interventions on the importance of gender and race features in influencing various risk factors are outlined in **Figure S.1** and **Figure S.2**, reinforcing our broader efforts toward model fairness across multiple sensitive attributes.

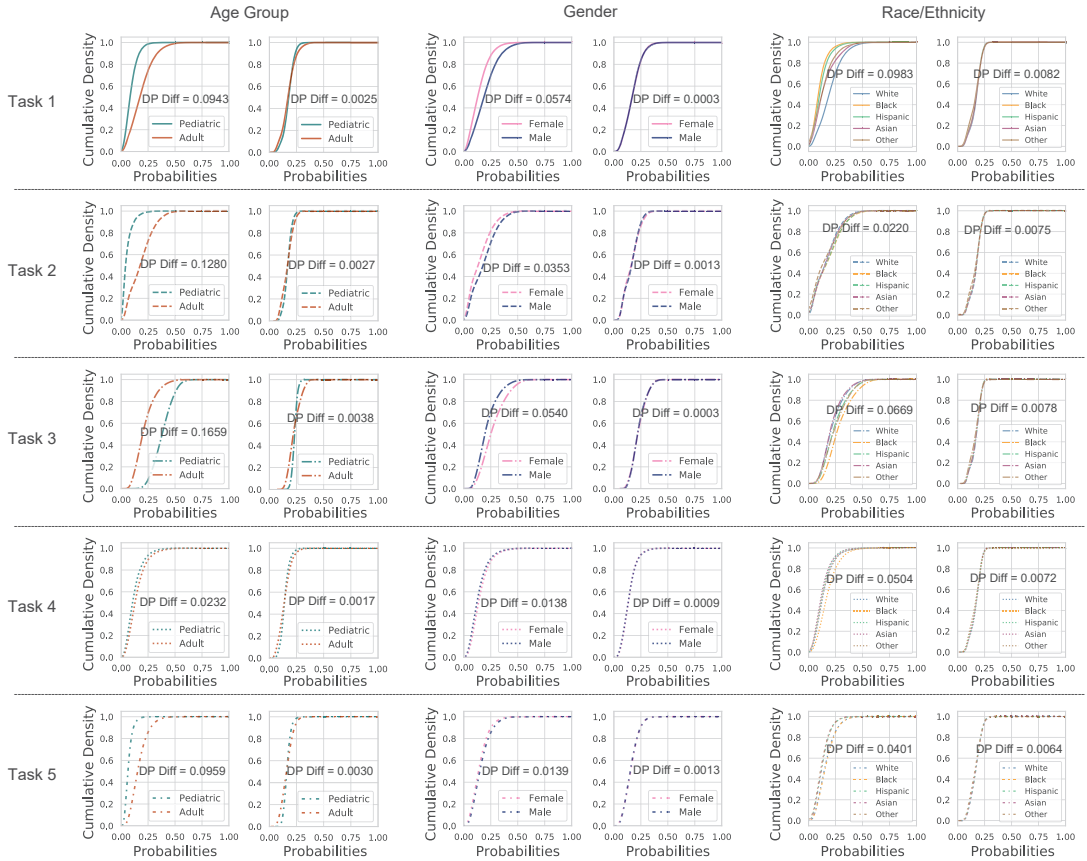

**Figure 5**. Comparison of Cumulative Density Functions (CDFs) for Predicted Probabilities Across Sensitive Attributes Using Multi-task AUROC-Weighted and fairness-achieved models. It represents the CDFs of predicted probabilities for five risks: malignancy (Task 1), diabetes (Task 2), rejection (Task 3), infection (Task 4), and cardiovascular complications (Task 5) in three sensitive attributes: age group, gender, and race/ethnicity. The left side displays the CDFs generated from the Multitask AUROC-Weighted model for each sensitive attribute setting. The right side showcases the CDFs derived from the proposed fairness-achieved algorithm. Within each setting, individual curves represent the specific subgroups of the corresponding sensitive attribute (e.g., male vs. female for gender, adult and pediatric for age group, White vs. Black vs. Hispanic vs. Asian, for etc., race/ethnicity).

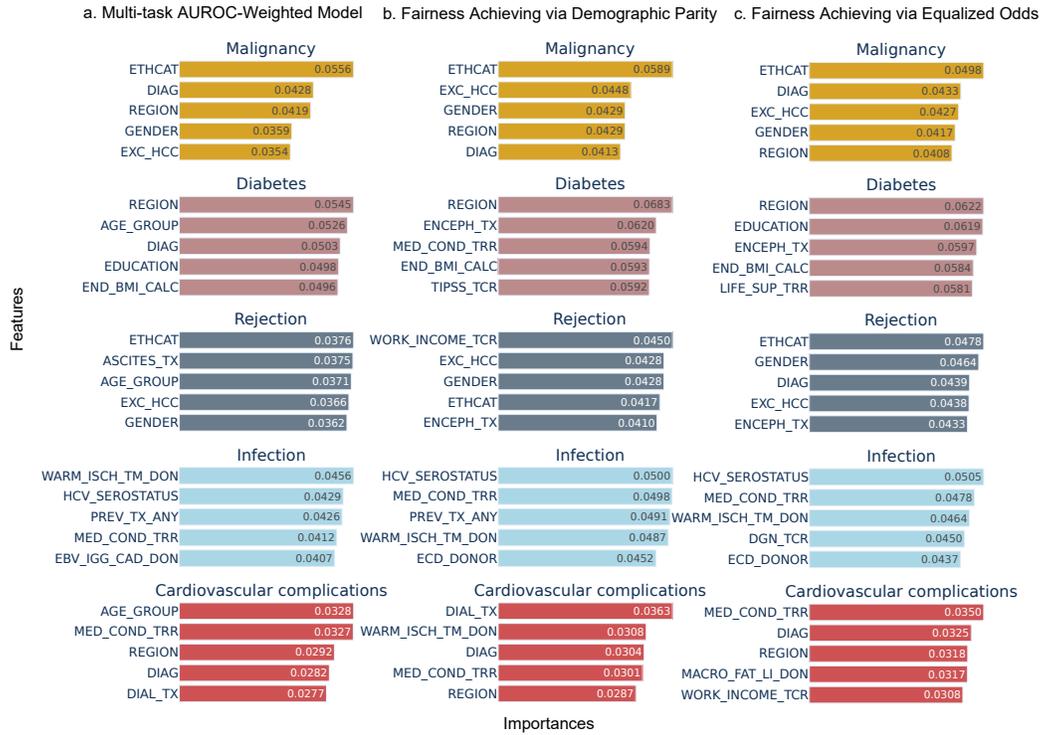

**Figure 6.** Comparative analysis of top 5 feature importance for post-liver transplant risk factors across different models. The top 5 feature importance for each post-liver transplant risk factor, determined by permutation feature importance, presented in three distinct settings: a) Multi-task AUROC-Weighted with task balancing. b) Model incorporating AUROC Weighted task balancing method and fairness-achieving method to promote Demographic Parity (DP), specifically aiming to mitigate the bias for age groups. c) Model integrating AUROC Weighted task balancing method and fairness-achieving method targeting Equalized Odds, which also focuses on the bias among age groups. The risk factors (from top to bottom) are malignancy, diabetes, rejection, infection, and cardiovascular complications. Each plot in the subfigures a, b, and c represents one of the settings described above.

## 6. Conclusion

In this study, we developed a multi-task learning model with task balancing to jointly predict multiple clinically important post-liver transplant risk factors. Our Transformer-based neural network model achieved accurate multi-task prediction and balanced performance across risk factors. Additionally, we proposed a fairness-achieving algorithm to ensure equal prediction performance across demographic subpopulations based on our proposed task-balancing model.

Our proposed balance-achieving and fairness-achieving algorithms enhance the prediction of post-liver transplant risk factors, potentially aiding the optimization of donor-patient matching in the existing liver transplant matching system. Our methods are also flexible and applicable to other domains. The choice of evaluation and fairness metrics is adaptable based on the specific needs and characteristics of the task. Consequently, our work demonstrates how multi-task learning, combined with task balancing and fairness measures, can elevate the prediction of multiple critical clinical risk factors.

**7. Discussion**

In conclusion, our multi-task learning model effectively incorporates task balancing and fairness-achieving algorithms, thus offering a promising framework for predicting multiple post-liver transplantation risk factors. Despite its robustness in balancing performance across tasks and mitigating demographic disparities, the model's efficacy is still subject to a few limitations.

First, the proposed method focuses on the model bias, specifically, the bias created during the training and inference of the prediction model itself. The model bias can be significantly reduced by designing new training strategies and loss functions. The performance and fairness of the resulting model can be sensitive to these hyperparameters, requiring additional effort for fine-tuning to achieve the best tradeoff between prediction accuracy and model fairness.

Additionally, similar to algorithms focused on achieving fairness, task-balancing algorithms often trade off the overall accuracy performance of the model to ensure equitable performance across multiple tasks. It poses a challenge in designing a well-optimized loss function and optimization technique that can find the best compromise among the three objectives: prediction accuracy, task balance, and fairness among different subgroups.

This study analyzed data from 1987 to 2018, highlighting complex changes in medical practices and patient demographics over time. Although we employed

Multiple Imputation by Chained Equations (MICE) for missing values and outliers [33], it may not capture temporal dynamics entirely. Through our primary focus are multiple risks prediction and fairness in liver transplantation, it is worth acknowledging the temporal dynamics and their potential influence. Future research should further explore demographics changes in liver transplant patients over past decades.

In conclusion, we have presented an innovative model and algorithm to tackle the aforementioned challenge. Our motivation arises from a critical observation in the field of liver transplantation, which necessitates accurate predictions and fairness. Apart from the binary outcome of transplant success or failure, multiple overlooked post-transplant risk factors are equal important. These factors collectively contribute to the transplant's overall success, influencing the patient's survival time and quality of life. Therefore, our study is the first attempt to develop a model that can predict multiple risks with comparable accuracy while ensuring fairness among different subgroups.

**Acknowledgment**

XJ is CPRIT Scholar in Cancer Research (RR180012), and he was supported in part by Christopher Sarofim Family Professorship, UT Stars award, UTHealth startup, the National Institute of Health (NIH) under award number R01AG066749, R01AG066749-03S1, R01LM013712, U24LM013755 and U01TR002062, and the National Science Foundation (NSF) #2124789. This work was supported in part by Health Resources and Services Administration contract HHSH250-2019-00001C. The content is the responsibility of the authors alone and does not necessarily reflect the views or policies of the Department of Health and Human Services, nor does mention of trade names, commercial products, or organizations imply endorsement by the U.S. Government.

# Supplementary Materials

**Table S.1.** Detailed Descriptions and Categories of Patient Features

| Features | Descriptions |
|---|---|
| **ABO** | Recipient Blood Group at Registration |
| **AGE_GROUP** | Recipient Age Group A=Adult P=Peds |
| **ALBUMIN_TX** | Recipient Serum Albumin at Transplant |
| **AMIS** | A Locus Mismatch Level |
| **ARTIFICIAL_LI_TCR** | Recipient On Artificial Liver At Listing |
| **ASCITES_TX** | Recipient Ascites at Transplant |
| **BACT_PERIT_TCR** | Recipient Spontaneous Bacterial Peritonitis at Registration |
| **BMIS** | B Locus Mismatch Level |
| **CMV_IGM** | Recipient-Cmv By Igm Test Result at Transplant |
| **CMV_STATUS** | Recipient Cmv Status at Transplant |
| **CREAT_TX** | Recipient Serum Creatinine At Time Of Tx |
| **DAYSWAIT_CHRON** | Days On Liver Waiting List |
| **DGN_TCR** | Primary Diagnosis At Time Of Listing |
| **DGN2_TCR** | Secondary Diagnosis At Time Of Listing |
| **DIAG** | Recipient Primary Diagnosis |
| **DIAL_TX** | Dialysis Prior Week To Transplant? |
| **DRMIS** | Dr Locus Mismatch Level |
| **EDUCATION** | Recipient Highest Educational Level at Registration |
| **ENCEPH_TX** | Encephalopathy At Transplant |
| **END_BMI_CALC** | Calculated Candidate Bmi At Removal/Current Time |
| **ETHCAT** | Recipient Ethnicity Category |
| **ETHNICITY** | Recipient Ethnicity |
| **EXC_HCC** | Type Of Exception Relative To Hcc: Hbl, Hcc,Non-Hcc: Hbl=Hepatoblastoma |

| Field | Description |
|---|---|
| **FINAL_MELD_PELD_LAB_SCORE** | Most Recent Waiting List Meld/Peld Lab Score Or At Removal If Removed(May Be Different Than Allocation Score/Endstat) |
| **FINAL_SERUM_SODIUM** | Most Recent Waiting List Serum Sodium Or At Removal If Removed |
| **FUNC_STAT_TCR** | Recipient Functional Status at Registration |
| **GENDER** | Recipient Gender |
| **HBEAB_OLD** | Recipient- Hepatitis B E Antibody Serology (Pre 10/25/99 Data) at Transplant |
| **HBV_CORE** | Recipient Hepatitis B-Core Antibody |
| **HBV_NAT** | Trr Hbv Nat Result |
| **HBV_SUR_ANTIGEN** | Recipient Hep B Surface Antigen |
| **HBV_SURF_TOTAL** | Recipient Hbv Surface Antibody Total at Transplant |
| **HCC_DIAG** | Has Recipient Ever Had A Diagnosis Of Hcc? |
| **HCV_SEROSTATUS** | Recipient Hep C Status |
| **HGT_CM_CALC** | Calculated Recipient Height(Cm) |
| **HIV_SEROSTATUS** | Recipient Hiv Serostatus At Transplant |
| **HLAMIS** | Hla Mismatch Level |
| **INR_TX** | Recipient Inr At Transplant |
| **LIFE_SUP_TRR** | Recipient Life Support Pre-Transplant at Transplant |
| **LIST_MELD** | Patient Listed Prior To Meld/Peld? |
| **MED_COND_TRR** | Recipient Medical Condition Pre-Transplant at Transplant |
| **PORTAL_VEIN_TCR** | Recipient History Of Portal Vein Thrombosis at Registration |
| **PORTAL_VEIN_TRR** | Recipient Portal Vein Thrombosis at Transplant |
| **PREV_AB_SURG_TCR** | Recipient Previous Upper Abdominal Surgery at Registration |
| **PREV_AB_SURG_TRR** | Recipient Previous Upper Abdominal Surgery at Transplant |
| **PREV_TX_ANY** | Calculated Previous Transplant Of Any Organ Type |
| **PRVTXDIF** | Recipient Days Between Previous And Current Transplant |
| **REGION** | Wl Unos/Optn Region Where Listed/Transplanted |
| **TBILI_TX** | Recipient Total Bilirubin at Transplant |
| **TIPSS_TCR** | Recipient Transjugular Intrahepatic Portacaval Stint Shunt(Tipss) at Registration |
| **VENTILATOR_TCR** | Recipient On Life Support - Ventilator at Registration (1=Yes, 0=No) |
| **WORK_INCOME_TCR** | Work For Income At Registration? |

**Table S.2.** Detailed Descriptions and Categories of Donor Features

| Features | Descriptions |
|---|---|
| **ABO_DON** | Donor Blood Type |
| **ABO_MAT** | Donor-Recipient Abo Match Level |
| **AGE_DON** | Donor Age (Yrs) |
| **ALCOHOL_HEAVY_DON** | Ddr Heavy Alcohol Use (Y/N/U) |
| **ANTIHYPE_DON** | Deceased Donor-Antihypertensives W/In 24 Hrs Pre-Cross Clamp |
| **ARGININE_DON** | Deceased Donor-Was Donor Given Arginine Vasopressin Within 24 Hrs Pre Cross Clamp? |
| **BLOOD_INF_DON** | Deceased Donor-Blood As Infection Source |
| **BUN_DON** | Deceased Donor-Terminal Blood Urea Nitrogen |
| **CARDARREST_NEURO** | Deceased Donor-Cardiac Arrest Post Brain Death |
| **CDC_RISK_HIV_DON** | Deceased Donor - Per Phs, Does The Donor Have Risk Factors For Blood-Borne Disease Transmission? |
| **CMV_DON** | Donor Serology Anti Cmv (For Living Donor, Pre Unet Data Only) |
| **COD_CAD_DON** | Deceased Donor-Cause Of Death |
| **CONTIN_CIG_DON** | Deceased Donor-History Of Cigarettes In Pastat >20Pack Yrs+Recent 6Mo Use |
| **CONTIN_COCAINE_DON** | Deceased Donor-History Of Cocaine Use+Recent 6Mo Use |
| **CONTIN_OTH_DRUG_DON** | Deceased Donor-History Of Other Drugs In Past+Recent 6Mo Use |
| **CONTROLLED_DON** | Deceased Donor Controlled Non-Heart Beating (Y/N) |
| **CREAT_DON** | Deceased Donor-Terminal Lab Creatinine |
| **DDAVP_DON** | Deceased Donor-Synthetic Anti Diuretic Hormone (Ddavp) |
| **DEATH_CIRCUM_DON** | Deceased Donor-Circumstance Of Death |
| **DEATH_MECH_DON** | Deceased Donor-Mechanism Of Death |
| **DIABDUR_DON** | Deceased Donor-Diabetes Duration |
| **DIABETES_DON** | Deceased Donor-History Of Diabetes (Y,N) |

| Field | Description |
|---|---|
| **DIET_DON** | Deceased Donor-Hypertension Diet Controlled |
| **DONCRIT_MAX_AGE** | Donor Criteria - Maximum Acceptable Donor Age (Months) - Local |
| **DONCRIT_MAX_MILE** | Donor Criteria - Maximum Miles The Organ Or Recovery Team Will Travel |
| **DONCRIT_MAX_WGT** | Donor Criteria - Maximum Acceptable Donor Weight (Kg) - Local |
| **DONCRIT_MIN_WGT** | Donor Criteria - Minimum Acceptable Donor Weight (Kg) - Local |
| **EBV_IGG_CAD_DON** | Deceased Donor Epstein Barr Virus By Igg Test Result |
| **EBV_IGM_CAD_DON** | Deceased Donor Epstein Barr Virus By Igm Test Result |
| **ECD_DONOR** | Expanded Donor Per Kidney Allocation Definition 1=Yes |
| **HBSAB_DON** | Deceased Donor Hbsab Test Result |
| **HBV_CORE_DON** | Donor Hbv Core Antibody |
| **HBV_NAT_DON** | Ddr Hbv Nat Results: |
| **HEMATOCRIT_DON** | Ddr:Hematocrit: |
| **HEP_C_ANTI_DON** | Deceased Donor-Antibody To Hep C Virus Result |
| **HGT_CM_DON_CALC** | Calculated Donor Height (Cm) |
| **HIST_CANCER_DON** | Deceased Donor-History Of Cancer (Y/N) |
| **HIST_CIG_DON** | Deceased Donor-History Of Cigarettes In Past at >20Pack Yrs |
| **HIST_COCAINE_DON** | Deceased Donor-History Of Cocaine Use In Past |
| **HIST_HYPERTENS_DON** | Deceased Donor-History Of Hypertension |
| **HIST_INSULIN_DEP_DON** | Deceased Donor-Insulin Dependent Diabetes (Y,N) |
| **HIST_OTH_DRUG_DON** | Deceased Donor-History Of Other Drug Use In Past |
| **HISTORY_MI_DON** | Ddr History Of Prior Mi |
| **INSULIN_DON** | Deceased Donor-Was Donor Given Insulin Within 24 Hrs Pre Cross Clamp? |
| **LI_BIOPSY** | Cadaver Donor Liver Biopsy |
| **MACRO_FAT_LI_DON** | Deceased Donor Macro Fat% |

| Field | Description |
|---|---|
| **MICRO_FAT_LI_DON** | Deceased Donor Micro Fat% |
| **NON_HRT_DON** | Deceased Donor-Non-Heart Beating Donor |
| **PH_DON** | Ddr:Blood Ph: |
| **PROTEIN_URINE** | Deceased Donor Protein In Urine |
| **PT_DIURETICS_DON** | Deceased Donor-Diuretics B/N Brain Death W/In 24 Hrs Of Procurement |
| **PT_STEROIDS_DON** | Deceased Donor-Steroids B/N Brain Death W/In 24 Hrs Of Procurement |
| **PT_T3_DON** | Deceased Donor-Triiodothyronine-T3 B/N Brain Death W/In 24 Hrs Of Procurement |
| **PT_T4_DON** | Deceased Donor-Thyroxine-T4 B/N Brain Death W/In 24 Hrs Of Procurement |
| **PULM_INF_DON** | Deceased Donor-Infection Pulmonary Source |
| **RESUSCIT_DUR** | Deceased Donor-Resuscitation Following Cardiac Arrest Post Brain Death |
| **SGOT_DON** | Deceased Donor-Terminal Sgot/Ast |
| **SGPT_DON** | Deceased Donor-Terminal Sgpt/Alt |
| **SKIN_CANCER_DON** | Deceased Donor-Skin Cancer At Procurement (Y/N) |
| **TBILI_DON** | Deceased Donor-Terminal Total Bilirubin |
| **TRANSFUS_TERM_DON** | Ddr:Number Of Transfusions During This (Terminal) Hospitalization: |
| **URINE_INF_DON** | Deceased Donor-Infection Urine Source |
| **VASODIL_DON** | Deceased Donor-Vasodilators W/In 24Hrs Pre-Cross Clamp |
| **VDRL_DON** | Deceased Donor-Rpr-Vdrl Result |
| **WARM_ISCH_TM_DON** | Deceased Donor Non-Heart Beating Estimated Warm Ischemic Time (Minutes) |

**Figure S.1.** Comparative analysis of top 5 feature importance for post-liver transplant risk factors across different models. The top 5 feature importance for each post-liver transplant risk factor, determined by permutation feature importance, presented in three distinct settings: a) Multi-task AUROC-Weighted without fairness-achieving. b) Model incorporating AUROC Weighted task balancing method and fairness-achieving method to promote Demographic Parity (DP), specifically aiming to mitigate gender bias. c) Model integrating AUROC Weighted task balancing method and fairness-achieving method targeting Equalized Odds, which also focuses on the bias among genders. The risk factors (from top to bottom) are malignancy, diabetes, rejection, infection, and cardiovascular complications. Each plot in the subfigures a, b, and c represents one of the settings described above.

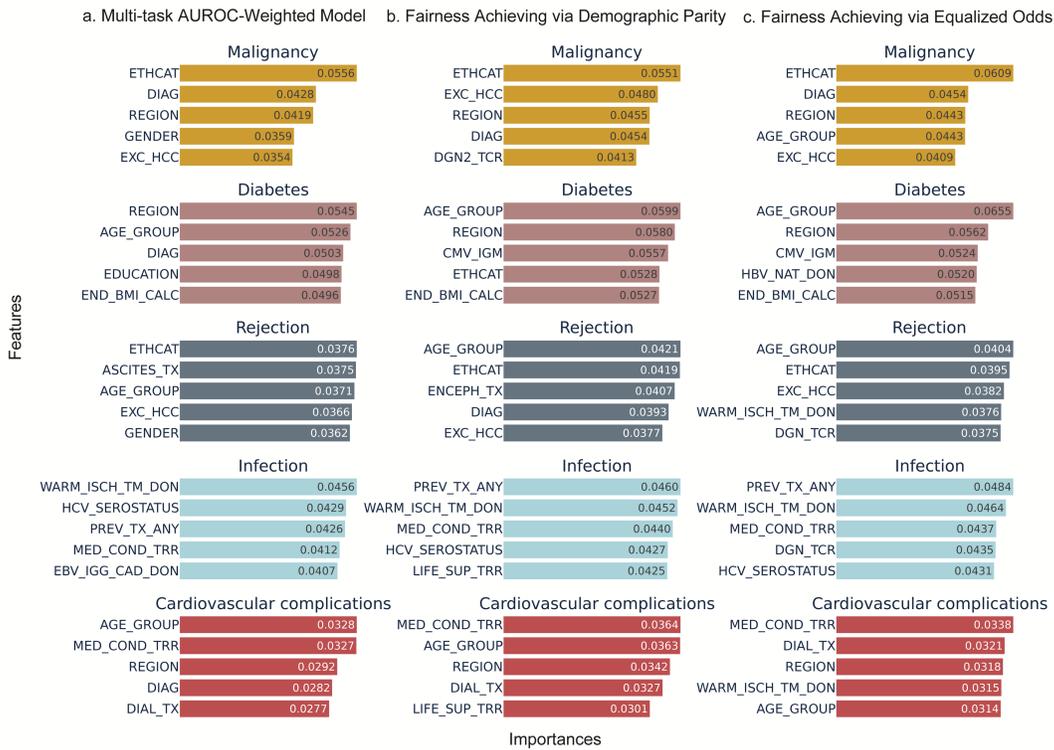

**Figure S.2.** Comparative analysis of top 5 feature importance for post-liver transplant risk factors across different models. The top 5 feature importance for each post-liver transplant risk factor, determined by permutation feature importance, presented in three distinct settings: a) Multi-task AUROC-Weighted without fairness-achieving. b) Model incorporating AUROC Weighted task balancing method and fairness-achieving method to promote Demographic Parity (DP), specifically aiming to mitigate the bias for race/ethnicity. c) Model integrating AUROC Weighted task balancing method and fairness-achieving method targeting Equalized Odds, which also focuses on the bias among race/ethnicity. The risk factors (from top to bottom) are malignancy, diabetes, rejection, infection, and cardiovascular complications. Each plot in the subfigures a, b, and c represents one of the settings described above.

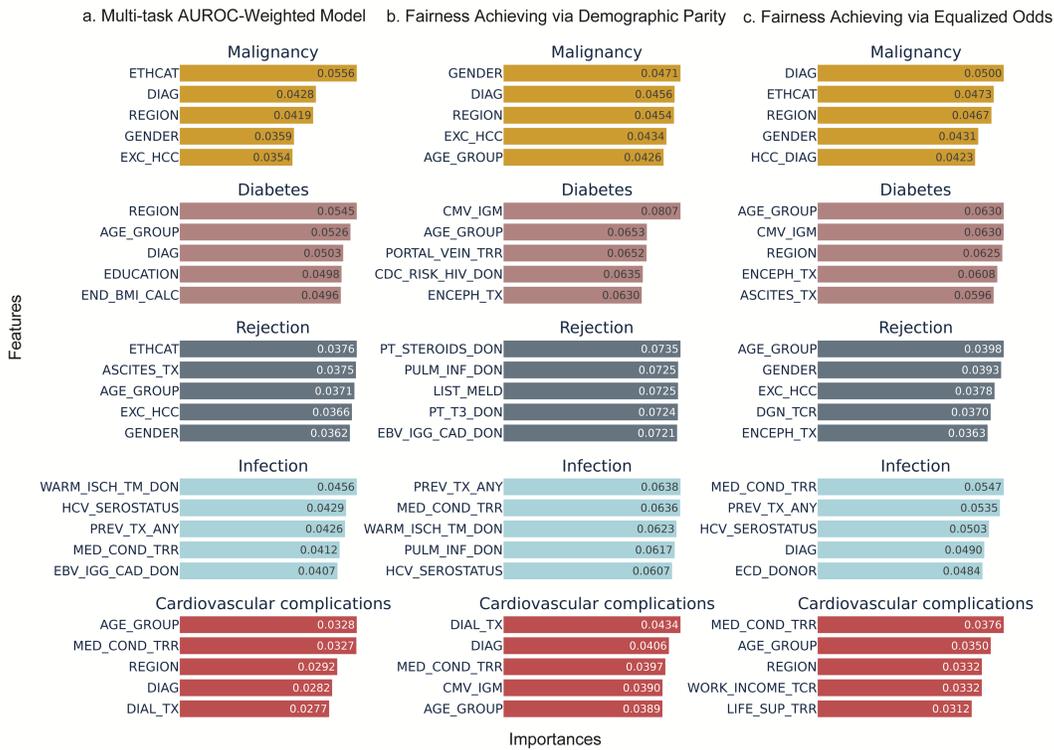